\documentclass[letterpaper, 10 pt, conference]{ieeeconf}  

\IEEEoverridecommandlockouts                              

\usepackage{graphicx}
\usepackage{afterpage}
\usepackage{color}
\usepackage{siunitx}
\usepackage{changepage}
\usepackage{mathrsfs}
\usepackage[normalem]{ulem}
\usepackage{color}

\usepackage{bm,amsmath,amsfonts,amssymb,dsfont} 

\usepackage{hyperref}
\definecolor{mydarkblue}{rgb}{0,0.08,0.45}
\hypersetup{
    colorlinks=false, 
    linkcolor=mydarkblue,
    citecolor=mydarkblue,
    filecolor=mydarkblue,
    urlcolor=mydarkblue,
}

\def\Del#1{{\color{red}\sout{#1}}}

\def\Fig#1{Fig.~\ref{#1}}
\def\Tab#1{Table~\ref{#1}}
\def\Sec#1{Section~\ref{#1}}

\newcommand{\cmmnt}[1]{\ignorespaces}
\newcommand{\mat}[1]{{\ensuremath{{\mathbf{#1}}}}}

\newcommand{\NewR}{\ensuremath{\mathds{R}}}

\title{\LARGE \bf
Large-scale 6D Object Pose Estimation Dataset for Industrial Bin-Picking
}

\author{Kilian Kleeberger$^{1}$, Christian Landgraf$^{1}$, and Marco F. Huber$^{1,2}$
\thanks{$^{1}$Kilian Kleeberger and Christian Landgraf are with the Department Robot and Assistive Systems and Marco Huber is with the Center for Cyber Cognitive Intelligence (CCI), Fraunhofer Institute for Manufacturing Engineering and Automation IPA, Nobelstra{\ss}e 12, 70569 Stuttgart, Germany
{\tt\small \{kilian.kleeberger, christian.landgraf, marco.huber\}@ipa.fraunhofer.de}}%
\thanks{$^{2}$Marco F. Huber is with the Institute of Industrial Manufacturing and Management IFF, University of Stuttgart, Allmandring 35, 70569 Stuttgart, Germany
{\tt\small marco.huber@ieee.org}}%
}

\begin{document}

\maketitle
\thispagestyle{empty}
\pagestyle{empty}

\begin{abstract}
In this paper, we introduce a new public dataset for 6D object pose estimation and instance segmentation for industrial bin-picking. The dataset comprises both synthetic and real-world scenes. For both, point clouds, depth images, and annotations comprising the 6D pose (position and orientation), a visibility score, and a segmentation mask for each object are provided. Along with the raw data, a method for precisely annotating real-world scenes is proposed.

To the best of our knowledge, this is the first public dataset for 6D object pose estimation and instance segmentation for bin-picking containing sufficiently annotated data for learning-based approaches. Furthermore, it is one of the largest public datasets for object pose estimation in general. The dataset is publicly available at \url{http://www.bin-picking.ai/en/dataset.html}.
\end{abstract}

\section{Introduction}

\begin{figure}[tb]
\centering
\includegraphics[scale=0.89]{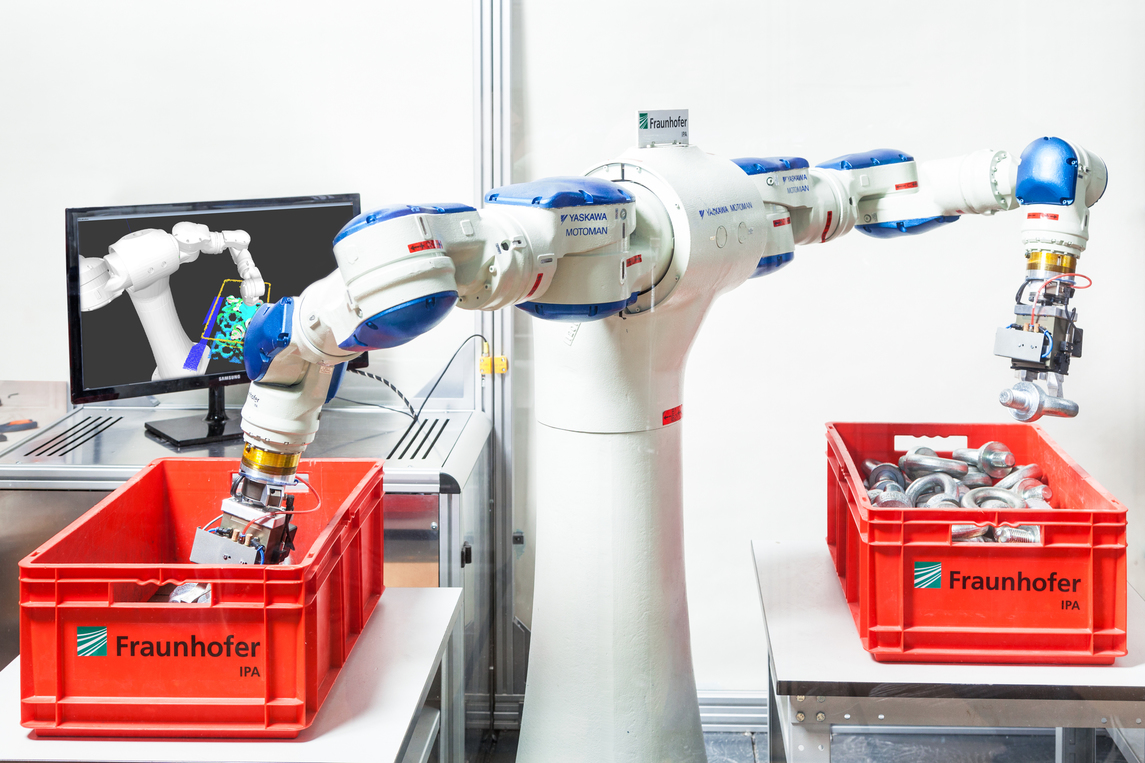}
\caption{An experimental real-world \href{https://www.youtube.com/watch?v=xhTkgajg8wQ}{robot cell} from Fraunhofer IPA. A dual arm robot takes ring screws out of a bin and drops them in another one. A 3D sensor is mounted above each bin.}
\label{fig:robotcell}
\end{figure}

The bin-picking application is concerned with a robot that has to grasp single instances of rigid objects from a chaotically filled bin (see \Fig{fig:robotcell}). In the context of industrial productions, it aims for replacing manual extraction out of storage boxes.
In this case, object pose estimation is a challenging task due to cluttered scenes with heavy occlusions and many \cmmnt{similar-looking distractors} identical
objects. Typically, one is forced to predict 6D poses from a single depth image or point cloud.

Since the beginning of the deep learning era, the performance on
many computer vision tasks increased drastically. This is closely related to the availability of large real-world datasets as for image classification~\cite{imagenet2009}, object detection and segmentation~\cite{coco2014}, or autonomous driving~\cite{kitti2013, cityscapes2016}. However, annotating vast amounts of real-world data is time-consuming and tedious, especially for 3D data.

Previous approaches towards datasets for 6D object pose estimation such as~\cite{linemod2012, ic_tejani2014, tless2017, sileane2017} either lack in the amount of data, the variety of scenes or do not fit the bin-picking scenario. According to the experiments in~\cite{bop2018}, the currently leading method is based on point-pair features, which is in contrast to many other computer vision tasks being dominated by approaches based on deep learning.
\cmmnt{\Del{This contribution aims to support learning-based methods and to leverage object pose estimation in the bin-picking scenario.}}

This paper aims for supporting machine learning methods and leveraging 6D object pose estimation in bin-picking scenarios. For this purpose, we provide a new large-scale benchmark dataset referred as ``Fraunhofer IPA Bin-Picking dataset'' including 520 fully annotated point clouds and corresponding depth images of real-world scenes and about 206,000 synthetic scenes. It comprises eight objects from~\cite{sileane2017} as well as two newly introduced ones. We present an effective way to semi-automatically produce ground truth data of real-world bin-picking scenes using the iterative closest point (ICP) algorithm and a reconstruction of the scenes in a physics simulation. This comprises not only a translation vector $\mat{t} \in \NewR^3$ and a rotation matrix ${\mat R \in \mathrm{SO}(3)}$, but also a visibility score and a segmentation mask for each object. The synthetic data consists of approximately 198,000 annotated scenes for training and 8,000 scenes for testing. In contrast to most of other public datasets for 6D object pose estimation, our contribution can be used additionally for instance segmentation and contains a visibility score, being of great value in bin-picking scenarios.

This paper is structured as follows. The next section reviews related work. The novel 6D object pose estimation dataset is described in \Sec{sec:dataset}. \Sec{sec:evaluation}
gives information regarding an evaluation.
The paper closes with conclusions and an outlook to future work.


\section{Related Work}
\label{sec:related-work}
\begin{table*}[ht]$ $
\caption{Overview of related datasets for 6D object pose estimation and our contribution.
Pose and data independence implies no connection between different scenes, e.g., the same scene is not recorded from different viewpoints.}
\label{table:datasets}
\resizebox{\textwidth}{!}{%
\begin{tabular}{|l||c|c|c|c|c|c|c|c|c|c|}
\hline
dataset & data & no. of & $\approx$ \#real-world/ & (real-world) & instance & top & pose and data & clutter/ & visibility & homo- \\ 
& modality & objects & \#synthetic & annotation & segmentation & view & independence & occlusion & score & geneous \\
\hline \hline
\href{http://staffhome.ecm.uwa.edu.au/~00053650/recognition.html}{Mian et al.}~\cite{mian2006} & Point cloud & 5 & 50/- & manual & no & no & yes & yes/yes & yes & no\\ 
\hline
\href{http://campar.in.tum.de/Main/StefanHinterstoisser}{LINEMOD}~\cite{linemod2012} & RGB-D & 15 & 18,000/20,000 & rel. marker & no & no & no & yes/lim. & no & no \\
\hline
\href{https://hci.iwr.uni-heidelberg.de/vislearn/iccv2015-occlusion-challenge/}{Occlusion LINEMOD}~\cite{linemod_occ2014} & RGB-D & 8 & 1,214/- & rel. marker & no & no & no & yes/yes & no & no \\
\hline
\href{https://sites.google.com/site/ujwalbonde/publications/downloads}{Desk3D}~\cite{desk3d2014} & Point cloud & 6 & 850/- & rel. marker & no & no & no & yes/yes & no & no \\
\hline
\href{https://rse-lab.cs.washington.edu/projects/posecnn/}{YCB-Video}~\cite{PoseCNN} & RGB-D & 21 & 134,000/- & semi-auto. & no & no & no & yes/yes & no & no \\
\hline
\href{http://rkouskou.gitlab.io/research/LCHF.html}{Tejani et al.}~\cite{ic_tejani2014} & RGB-D & 6 & 5,000/- & rel. marker & no & no & no & yes/lim. & no & no \\ 
\hline
\href{http://rkouskou.gitlab.io/research/6D_NBV.html}{Doumanoglou et al.}~\cite{ic_doumanoglou2015} & RGB-D & 8 & 536/- & rel. marker & no & partly & no & yes/yes & no & partly \\ 
\hline
\href{http://www.pracsyslab.org/rutgers_apc_rgbd_dataset}{Rutgers APC}~\cite{rutgers2016} & RGB-D & 24 & 10,000/- & (semi) manual & no & no & no & yes/yes & no & no \\ 
\hline
\href{http://cmp.felk.cvut.cz/t-less/}{T-Less}~\cite{tless2017} & RGB-D & 30 & 48,000/77,000 & semi-auto. & no & no & no & yes/yes & no & no \\ 
\hline
\href{http://cmp.felk.cvut.cz/sixd/challenge_2017/}{BOP: TUD+TYO-L}~\cite{bop2018} & RGB-D & 24 & 64,000/56,000 & semi-auto. & partly & no & no & no/no & no & no \\ 
\hline
\href{https://rbregier.github.io/dataset2017}{Sil\'{e}ane}~\cite{sileane2017} & RGB-D & 8 & 679/1922 & obj. marker & yes & yes & yes & yes/yes & yes & yes \\
\hline
\href{http://www.bin-picking.ai/}{Fraunhofer IPA} & D./P.c./RGB & 10 & 520/206,000 & semi-auto. & yes & yes & synth. only & yes/yes & yes & yes\\
\hline
\end{tabular}
} 
\end{table*}

Particularly since the emergence of affordable sensors capable of recording 3D data, numerous corresponding datasets appeared as listed in~\cite{firman2016}. Nevertheless, the vast majority of them is unsuitable for machine learning methods for 6D object pose estimation due to missing ground truth information.
Mian et al.~\cite{mian2006} provided point clouds of scenes with different objects, but those contain neither a high amount of clutter nor multiple instances of the same object type.

The LINEMOD dataset~\cite{linemod2012} is a popular and commonly used benchmark containing about 18,000 RGB-D images of 15 texture-less objects. The work was augmented by~\cite{linemod_occ2014} such that ground truth poses are available for all objects depicted in the images. This enables the consideration of a higher degree of occlusion for evaluation.

Datasets sharing similar properties were presented in~\cite{desk3d2014, PoseCNN, ic_tejani2014, ic_doumanoglou2015}. All of these datasets have limited pose variability and data redundancy since only the very same scene is recorded from different angles (see \Tab{table:datasets}). Additionally, only Doumanoglou et al.~\cite{ic_doumanoglou2015} provided homogeneous scenes as it is the case in industrial bin-picking, i.e., multiple instances of the same object type are present in one image.

In the Rutgers APC dataset~\cite{rutgers2016}, a cluttered warehouse scenario with occlusion and 6,000 real-world test images of 24 objects is introduced, but merely includes non-rigid, textured objects and it is not targeted on bin-picking.

The T-Less~\cite{tless2017} dataset provides 38,000 real training images of 30 industrial texture-less objects plus 10,000 test images of 20 scenes by systematically sampling a sphere. Again, it lacks in homogeneity, has limited pose variability and exhibits data redundancy.

Due to the time-consuming and difficult process of annotating, most approaches use markers either on the object itself or relative to the objects to automatically produce ground truth data. The same scene is recorded multiple times causing data redundancy and pose inflexibility. However, after removing redundant scenes, the datasets become too small to be applicable for machine learning methods like deep neural networks.

BOP~\cite{bop2018} attempts to standardize and integrate the presented datasets in one novel benchmark for 6D object pose estimation. In addition, two new scenarios with varying lighting conditions were included, but those are not related to bin-picking. Similar to our approach, an ICP algorithm is used to refine the manually created ground truth. The work includes results from the SIXD Challenge 2017\footnote{\url{http://cmp.felk.cvut.cz/sixd/challenge_2017/}, last accessed on July 31, 2019.}, which focused on 6D object pose estimation of single instances of one object.

Furthermore, the common standard in industrial bin-picking uses top view 3D sensors, which is not the case in all aforementioned works. In the Sil\'{e}ane dataset~\cite{sileane2017}, this common practice is recognized and a
procedure to automatically annotate real-world images is presented.
However, this dataset provides at most 325 images of one object, which is usually far from being suitable to use advanced machine learning methods. In this work, we extend
the Sil\'{e}ane dataset to be large enough for learning-based methods and introduce two new industrial objects together with real-world data.

\section{Fraunhofer IPA Bin-Picking dataset}
\label{sec:dataset}

\begin{figure*}[ht]
\centering
\resizebox{0.95\linewidth}{!}{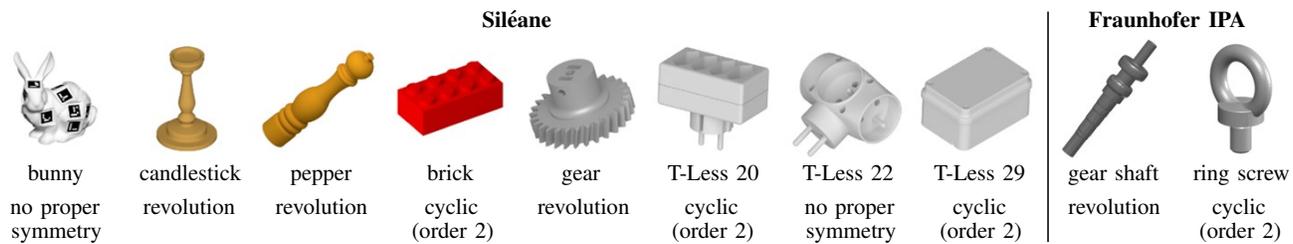}
\caption{An overview of the objects of the dataset and their proper symmetries based on~\cite{sileane2017}.}
\label{fig:workpieces}
\end{figure*}

In this section, we give details on the new dataset for 6D object pose estimation for industrial bin-picking.

\subsection{Sensor Setup}

\begin{table*}[hbtp]
\caption{Parameter setting of each object from our dataset.}
\label{table:workpieces}
\resizebox{\textwidth}{!}{%
\begin{tabular}{|l||c|c|c|c|c|c|c|c|}

\hline
object & resolution & clip start and & orthogonal & perspective & drop & number of & number of & number of \\ 
& & clip end & size & angle & limit & training cycles & test cycles & scenes \\
\hline \hline
Sil\'{e}ane Stanford bunny & $474 \times 506$ & $\approx$ 876--1,706 mm & 659 mm & \ang{35.3} & 80 & 250 & 10 & 21,060 \\ 
\hline
Sil\'{e}ane candlestick & $634 \times 618$ & $\approx$ 584--1,105 mm & 534 mm & \ang{36.0} & 60 & 250 & 10 & 15,860 \\ 
\hline
Sil\'{e}ane pepper & $506 \times 554$ & $\approx$ 920--1,606 mm & 724 mm & \ang{34.8} & 90 & 250 & 10 & 23,660 \\ 
\hline
Sil\'{e}ane brick & $506 \times 554$ & $\approx$ 878--1,031 mm & 117 mm & \ang{12.7} & 150 & 250 & 10 & 39,260 \\ 
\hline
Sil\'{e}ane gear & $1018 \times 1178$ & $\approx$ 1,639--2,082 mm & 478 mm & \ang{18.4} & 60 & 250 & 10 & 15,860\\ 
\hline
Sil\'{e}ane T-Less 20 & $634 \times 618$ & $\approx$ 584--1,105 mm & 534 mm & \ang{36.0} & 99 & 250 & 10 & 26,000 \\ 
\hline
Sil\'{e}ane T-Less 22 & $634 \times 618$ & $\approx$ 584--1,105 mm & 534 mm & \ang{36.0} & 100 & 250 & 10 & 26,260 \\
\hline
Sil\'{e}ane T-Less 29 & $634 \times 618$ & $\approx$ 584--1,105 mm & 534 mm & \ang{36.0} & 79 & 250 & 10 & 20,800 \\
\hline
Fraunhofer IPA gear shaft & $512 \times 512$ & 750--1,750 mm & 600 mm & \ang{22.0} & 30 & 250 & 10 & 8,060 \\ 
\hline
Fraunhofer IPA gear shaft (real-world) & $512 \times 512$ & 750--1,750 mm & 600 mm & \ang{22.0} & 22 & 0 & 10 & 230 \\
\hline
Fraunhofer IPA ring screw & $512 \times 512$ & 1,250--1,750 mm & 600 mm & \ang{20.0} & 35 & 250 & 10 & 9,360 \\ 
\hline
Fraunhofer IPA ring screw (real-world) & $512 \times 512$ & 1,250--1,750 mm & 600 mm & \ang{20.0} & 28 & 0 & 10 & 290 \\
\hline
\end{tabular}
} 
\end{table*}

We collected the real-world data using an Ensenso N20-1202-16-BL stereo camera having a minimum working distance of 1,000~mm, a maximum working distance of 2,400~mm, and an optimal working distance of 1,400~mm. The sensor produces images with a resolution of $1280 \times 1024$ pixels and is mounted above the bin. 

For the collection of synthetic data, we use
the same parameter settings in our physics simulation as in~\cite{sileane2017}. The detailed setting for each object for the clipping planes, the perspective angle of the perspective projection, the orthogonal size of the orthogonal projection, and image resolution are listed in \Tab{table:workpieces}.

\subsection{Dataset Description}
\label{subsec:datasetdescription}

We use eight objects with different symmetries from~\cite{sileane2017}, which again uses three objects originally published by~\cite{tless2017}. Moreover, we introduce two novel industrial objects: A gear shaft possessing a revolution symmetry and a ring screw possessing a cyclic symmetry. An overview is depicted in \Fig{fig:workpieces}.

The dataset is separated into a training and a test dataset. For each object in the test dataset, data is generated by iteratively filling an initially empty bin. This iterative procedure consists of ten cycles and ends if a particular drop limit is reached. For details of the iterative procedure see \Sec{sec:datacollection}. The training dataset comprises 250 cycles that are generated in the same way, but have no real-world scenes due to the time-consuming and non-scaling process of annotation.

However, the usage of sim-to-real transfer techniques such as domain randomization~\cite{dr2017} has proven to be successful and to allow high performance, as for example demonstrated by~\cite{dexthand2016}. In this way, the synthetic training dataset is sufficient to achieve high-quality results on the test set including the real-world dataset with techniques based on deep learning.

The synthetic scenes are independently filled, which means that there are no dependencies between the images. The number of objects in one scene is increased one by one, but after each scene is recorded, the bin is cleared and we start from scratch. In contrast, each real-world scene in one cycle depends on the previous scenes of this cycle due to our data collection procedure (\Sec{subsection:realworlddata}).

The ground truth data comprises a translation vector $\mat{t}$ and a rotation matrix $\mat{R}$ relative to the coordinate system of the 3D sensor, a visibility score $\mat{v} \in [0,1]$, and a segmentation image labeled by the object ID for perspective and orthogonal projection.


\begin{figure}[t]
\centering
\includegraphics[scale=0.36]{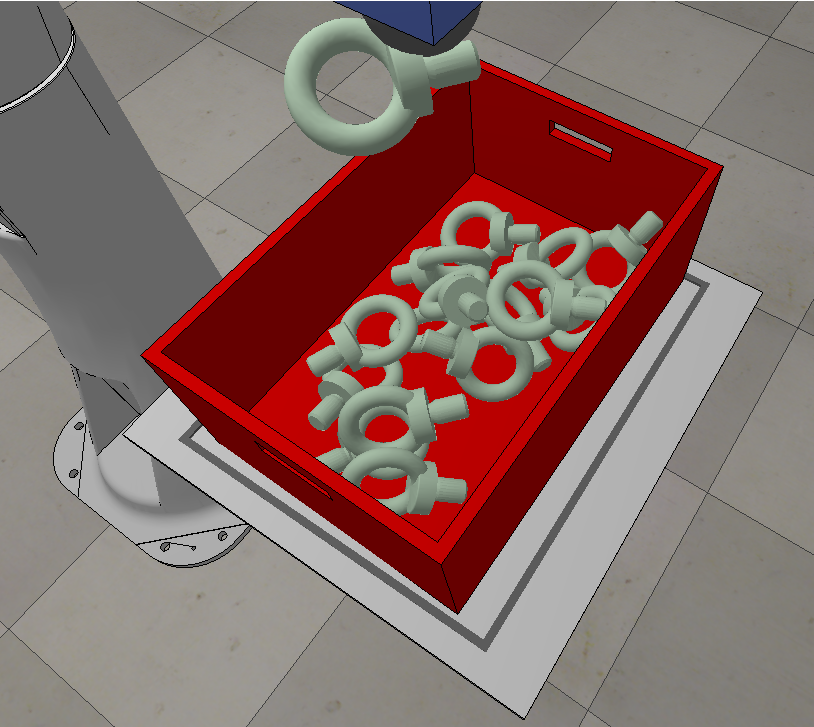}
\caption{Example scene of our physics simulation for data generation. New objects are dropped into a bin until a predefined drop limit is reached.}
\label{fig:vrep}
\end{figure}

\subsection{Data Collection Procedure} 
\label{sec:datacollection}

\begin{figure*}[p!]
\centering
\resizebox{0.79\linewidth}{!}{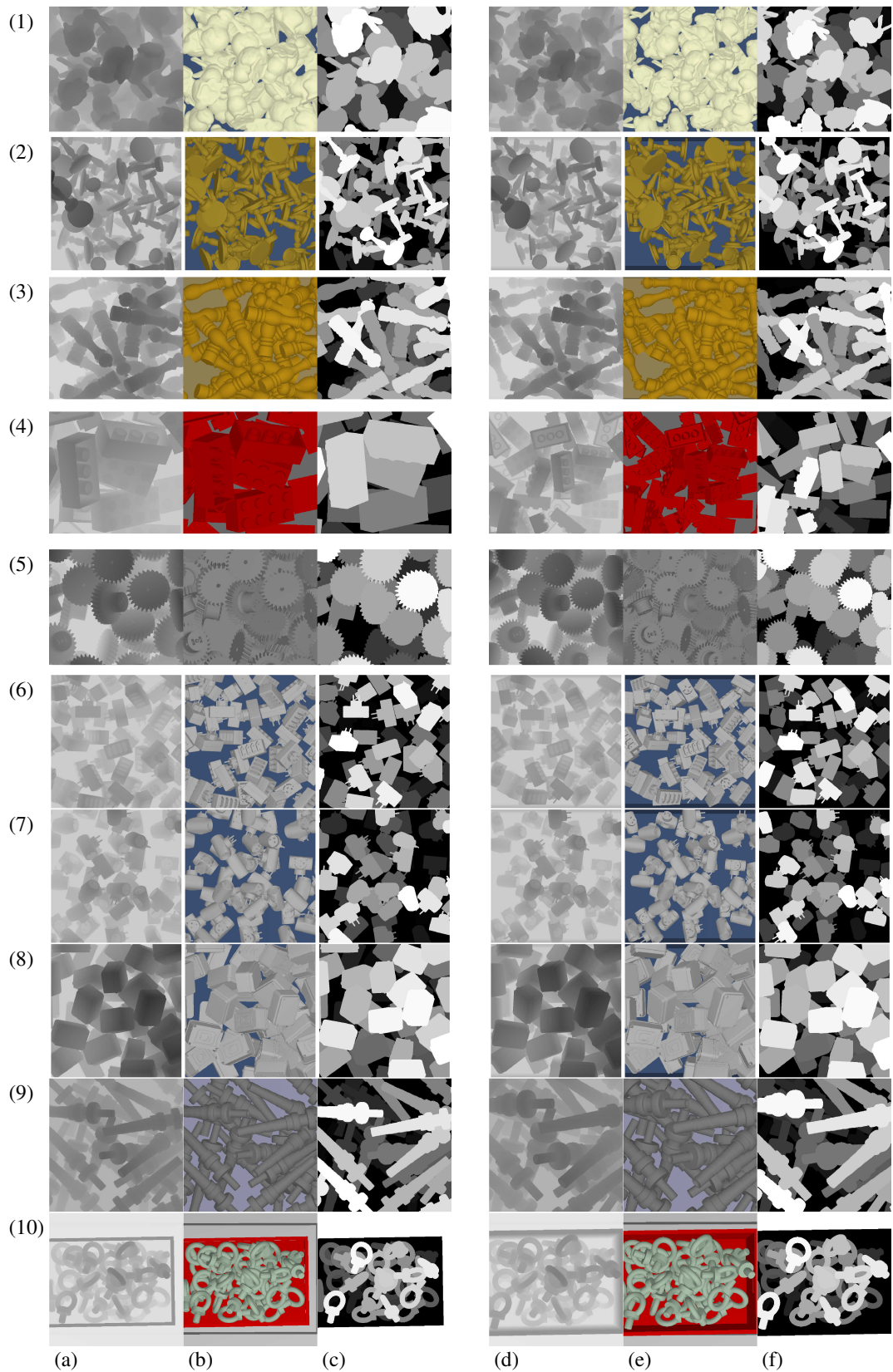}
\caption{Examples of synthesized images for each object: From left to right, each column gives examples for (a) orthogonal depth images, (b) orthogonal RGB images, (c) orthogonal segmentation images, (d) perspective depth images, (e) perspective RGB images, and (f) perspective segmentation images.
From top to bottom, row (1) shows Sil\'{e}ane Stanford bunnies, (2) Sil\'{e}ane candlesticks, (3) Sil\'{e}ane peppers, (4) Sil\'{e}ane bricks, (5) Sil\'{e}ane gears, (6) Sil\'{e}ane T-Less 20 objects, (7) Sil\'{e}ane T-Less 22 objects, (8) Sil\'{e}ane T-Less 29 objects, (9) Fraunhofer IPA gear shafts, and (10) Fraunhofer IPA ring screws.}
\label{fig:workpiecesexamples}
\end{figure*}

\begin{figure*}[ht]
\centering
\resizebox{0.9\linewidth}{!}{
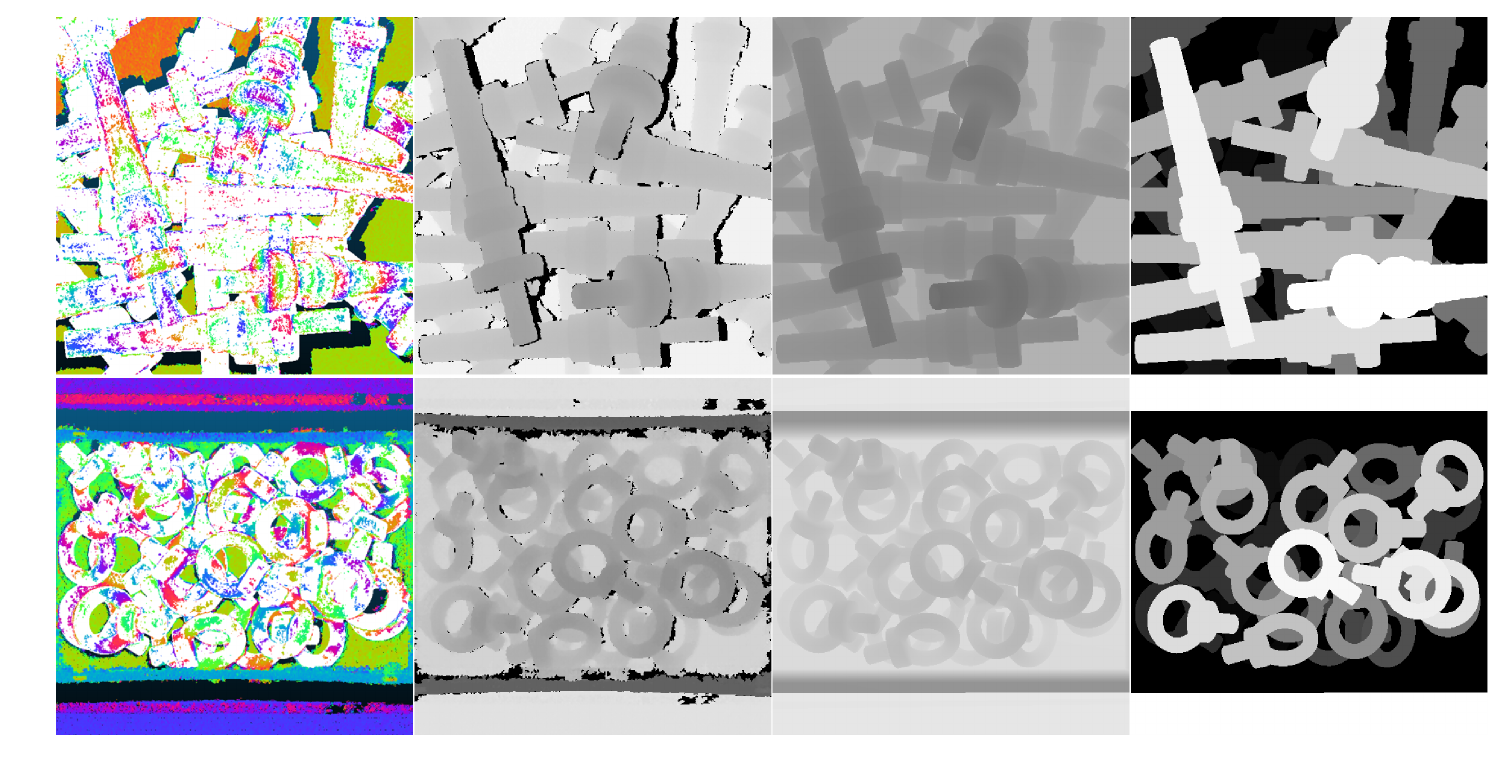
}
\caption{Examples of real-world data for Fraunhofer IPA gear shafts (1) and Fraunhofer IPA ring screws (2): (a) A real-world point cloud (colored) captured with an Ensenso N20-1202-16-BL stereo camera with ground truth object poses (white), (b) the real-world depth image, (c) the synthetic depth image of the reconstructed scene in simulation, and (d) the resulting segmentation image in perspective projection, respectively.}
\label{fig:realworldpointcloud}
\end{figure*}

\subsubsection{Synthetic Data}
\label{subsection:simulateddata}

To generate scenes typical for bin-picking, we use the physics simulation V-REP~\cite{vrep2013}. We import a CAD model of each object in the simulation and drop them from varying positions and with random orientation into the bin (see \Fig{fig:vrep}). To handle the dynamics and collisions, we use the built-in Bullet physics engine\footnote{\url{https://pybullet.org/wordpress/}, last accessed on July 31, 2019}. In favor of increasing realism for the newly introduced objects, we slightly shift the bin pose from image to image whereas the settings of the objects from~\cite{sileane2017} remain unchanged. Starting with an empty bin, we raise the number of objects dropped in the bin iteratively. This means, we drop one object in the first run, record the scene, clear the bin, drop two objects in the second run, etc. After each run, we record a depth image, a RGB image, and a segmentation image in both orthogonal and perspective projection together with the poses of all objects in the scene. This procedure is repeated until a predefined drop limit is reached and the collected data forms one cycle. \Fig{fig:workpiecesexamples} depicts example scenes.

The depth images are saved in 16 bit unsigned integer format (uint16).
The segmentation image is created by assigning the ID of the respective object to each pixel, i.e., zero for the bin, one for the first object, etc. If a pixel belongs to the background, the maximum value 255 of uint8 is assigned. For each item, we save a segmentation image containing only the individual object in order to calculate the total number of pixels forming this object. All other objects are made invisible for this single-object image. The final visibility score is calculated externally by the ratio between the visible pixels in the segmentation image and the total number of pixels.

If the object is partly outside of the original image, we further save a larger segmentation image containing the full object. The resolution of this image is increased to ensure the same number of pixels showing the original scene in the large image. Subsequently, the large image is used as reference to calculate the visibility score. The resulting value is computed for the orthogonal and perspective version and is attached to the aforementioned ground truth file containing the ID, class, translation vector, and rotation matrix of each object instance in the scene.


\subsubsection{Real-World Data}
\label{subsection:realworlddata}

Starting with a filled bin, we carefully remove the objects one by one and do not change the pose of the remaining ones. In each step, we record the 3D sensor data until the bin is empty. For annotation, we reverse the order of the scenes. For this purpose, we fit a point cloud representation of the object's CAD model to the newly added object by means of the ICP algorithm to get the precise 6D pose (see \Fig{fig:realworldpointcloud}).

With this result, we rebuild each real-world scene in our physics simulation and determine the segmentation mask and the visibility score for each object as described in \Sec{subsection:simulateddata}. We further provide images showing the merger of real-world depth images and the ground truth segmentation for each individual object to prove the quality of our annotation process.

\section{Evaluation}
\label{sec:evaluation}

Along with the dataset, we provide CAD models, Python tools for various conversion needs of point clouds and depth images or ground truth files and scripts to facilitate working with our dataset.


As demonstrated in~\cite{TM2019}, our synthetic dataset along with domain randomization~\cite{dr2017} can be used to get robust and accurate 6D pose estimates on our real-world scenes. By applying various augmentations to the synthetic images during training, the deep neural network is able to generalize to real-world data despite being entirely trained on synthetic data.

\subsection{Evaluation Metric} 


A common evaluation metric for 6D object pose estimation is ADD by Hinterstoisser et al.~\cite{linemod2012}, which accepts a pose hypothesis if the average distance of model points between the ground truth and estimated pose is less than 0.1 times the diameter of the smallest bounding sphere of the object. Because this metric cannot handle symmetric objects, ADI~\cite{linemod2012} was introduced for handling those. The ADI metric is widely used, but can fail to reject false positives as demonstrated in~\cite{sileane2017}. Therefore, we use the metric
provided by Br\'{e}gier et al.~\cite{sileane2018_distance, sileane2017}, which is suitable for rigid objects, for scenes of many parts in bulk, and properly considers cyclic and revolution object symmetries. A pose representative $\mat{p} \in \mathcal{R}(\mathcal{P})$ comprises a translation vector $\mat{t}$ and the relevant axis vectors of the rotation matrix $\mat R$ depending on the object's proper symmetry group.
The distance between a pair of poses $\mathcal{P}_1$ and $\mathcal{P}_2$ is defined as the minimum of the Euclidean distance between their respective pose representatives
\begin{equation*}
d(\mathcal{P}_1, \mathcal{P}_2) = \displaystyle\min\limits_{\mat{p}_1 \in \mathcal{R}(\mathcal{P}_1),~\mat{p}_2 \in \mathcal{R}(\mathcal{P}_2)} \lVert\mat{p}_1 - \mat{p}_2\rVert~.
\end{equation*}
A pose hypothesis is accepted (considered as true positive) if the minimum distance to the ground truth is less than 0.1 times the object's diameter. Following~\cite{sileane2017}, only the pose of objects that are less than 50\% occluded are relevant for the retrieval.
The metric breaks down the performance of a method to a single scalar value named average precision~(AP)
by taking the area under the precision-recall curve.

\subsection{Object Pose Estimation Challenge for Bin-Picking}
We are going to offer a competition for 6D object pose estimation for bin-picking at the IROS 2019. The proposed
dataset will serve as training and test dataset.
We hope to promote and facilitate the performance of industrial bin-picking robots with our contribution and advance the state-of-the-art for object pose estimation. Further information regarding the competition is available at \url{http://www.bin-picking.ai/en/competition.html}.

\section{Conclusions and Future Work}
To the best of our knowledge, we presented the first 6D object pose estimation benchmark dataset for industrial bin-picking allowing the usage of advanced machine learning methods.
It is composed of both synthetic and real-world scenes.
In future work, we plan to publish data of more objects.





\section*{Acknowledgment}
This work was partially supported by the Baden-W\"urttemberg Stiftung gGmbH (Deep Grasping -- Grant No. NEU016/1) and the Ministry of Economic Affairs of the state Baden-W\"urttemberg (Center for Cyber Cognitive Intelligence (CCI) -- Grant No. 017-192996). We would like to thank our colleagues for helpful discussions and comments.


\bibliographystyle{IEEEtran}
\bibliography{IEEEabrv,bibliography.bib}


\end{document}